\documentclass[10pt,twocolumn,letterpaper]{article}

\usepackage{iccv}
\usepackage{times}
\usepackage{epsfig}
\usepackage{graphicx}
\usepackage{amsmath}
\usepackage{amssymb}
\usepackage{algorithm2e}
\usepackage{booktabs}
\usepackage[toc,page]{appendix}
\newcommand{\ra}[1]{\renewcommand{\arraystretch}{#1}}

\iccvfinalcopy 

\def\iccvPaperID{2} 

\ificcvfinal\fi
\begin{document}

\title{Label and Sample: Efficient Training of Vehicle Object Detector from Sparsely Labeled Data}

\author{Xinlei Pan\\
UC Berkeley\\
Berkeley, CA, USA 94720\\
{\tt\small xinleipan@berkeley.edu}
\and
Sung-Li Chiang\\
UC Berkeley\\
Berkeley, CA, USA 94720\\
{\tt\small slchiang@berkeley.edu}
\and
John Canny \\
UC Berkeley\\
Berkeley, CA, USA 94720\\
{\tt\small canny@berkeley.edu}
}

\maketitle

\begin{abstract}
   Self-driving vehicle vision systems must deal with 
   an extremely broad and challenging set of scenes. 
   They can potentially exploit an enormous amount of 
   training data collected from vehicles in the field, 
   but the volumes are too large to train offline naively.
   Not all training instances are equally valuable though,
   and importance sampling can be used to prioritize which
   training images to collect. This approach assumes that
   objects in images are labeled with high accuracy. 
   To generate accurate labels in the field, we exploit 
   the spatio-temporal coherence of vehicle video. We use 
   a near-to-far labeling strategy by first labeling large,
   close objects in the video, and tracking them back in 
   time to induce labels on small distant presentations of
   those objects. In this paper we demonstrate the feasibility
   of this approach in several steps. 
   First, we note that an optimal subset (relative to all
   the objects encountered and labeled) of labeled objects in
   images can be obtained by importance sampling using gradients
   of the recognition network. Next we show that these gradients
   can be approximated with very low error using the loss function,
   which is already available when the CNN is running inference.
   Then, we generalize these results 
   to objects in a larger scene using an object detection 
   system. Finally, we describe a self-labeling scheme using 
   object tracking. Objects are tracked back in time 
   (near-to-far) and labels of near objects are used 
   to check accuracy of those objects in the far field.
   We then evaluate the accuracy of models trained on 
   importance sampled data vs models trained
   on complete data.
\end{abstract}

\section{Introduction}

Autonomous driving is receiving enormous development 
effort with many companies predicting large-scale 
commercial deployment in 2-3 years \cite{practical_or}. 
One of the most important features of autonomous 
driving vehicles is the ability to interpret the 
surroundings and perform complex perception task 
such as the detection and recognition of lanes, 
roads, pedestrians, vehicles, and traffic signs 
\cite{DBLP:journals/corr/RomeraBA16}. 
Recently, the growth of Convolutional Neural Networks (CNNs), 
and large labeled data sets \cite{deng2009imagenet,
everingham2010pascal} have led to tremendous 
progress in object detection and recognition 
\cite{RCNN, fast_rcnn, faster_rcnn, 
ssd}. It is now possible to detect 
objects with high accuracy 
\cite{faster_rcnn, YOLO, ssd}.  

\begin{figure}[t]
\centering
\includegraphics[width=\linewidth, height=0.6\linewidth]{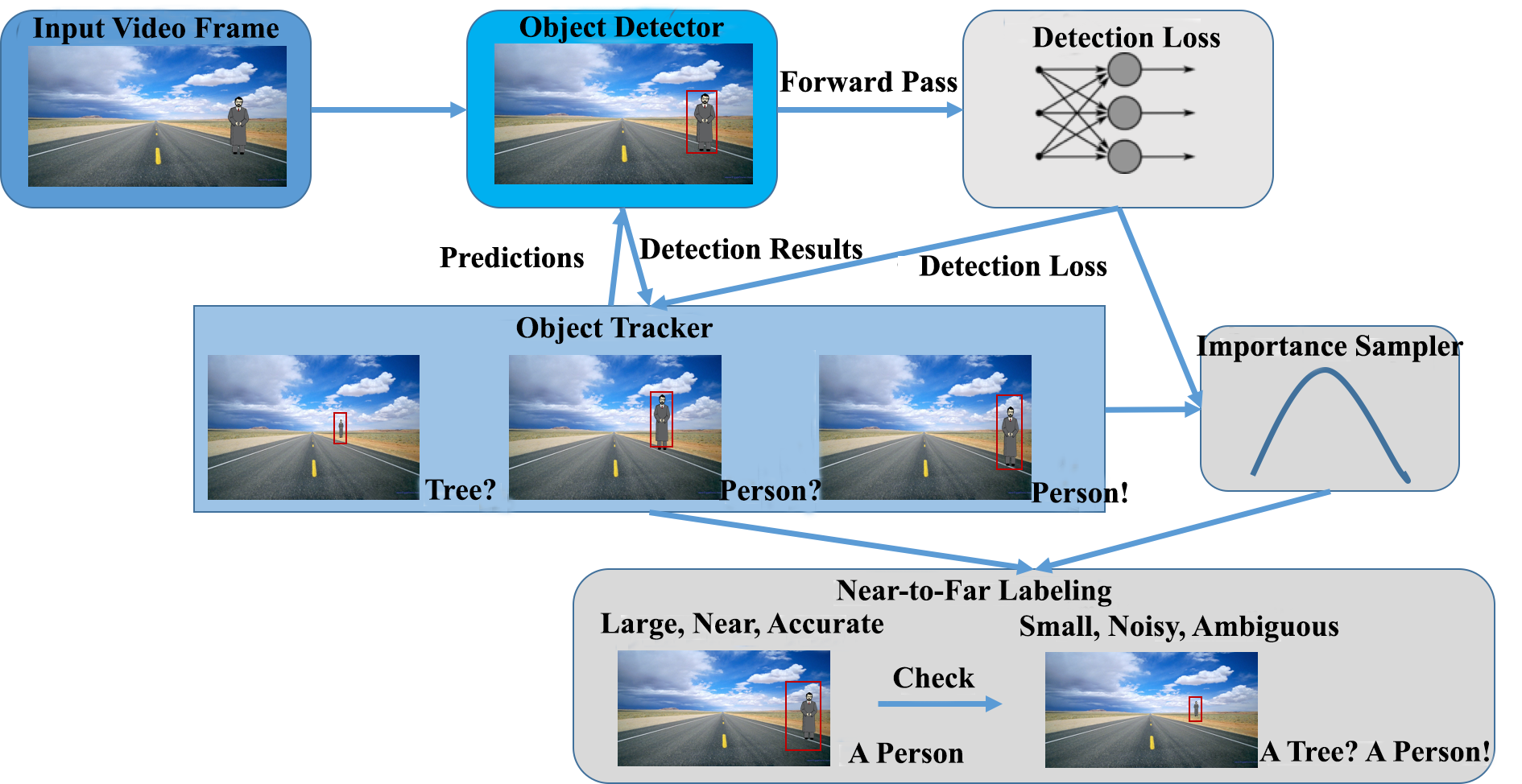}
\caption{Overview of our method. Given a sequence of 
video frame inputs, the object detection network first 
detects objects in a frame, the forward pass step. Then
the loss of the images are calculated. Both detection 
results and loss will be sent to object tracker 
which keeps a list of active objects. The importance 
sampler determines whether to save the detection or 
not based on the loss. Furthermore, the near-to-far 
labeling step checks the accuracy of objects that are 
in the far field using the classification result of 
near field objects, where we believe near field objects
are larger and the classification is more accurate.}
\label{fig1}
\end{figure}

Videos collected in-vehicle have a great potential 
to improve model quality (through offline training)
but at the scales achievable in a few years 
(billions to trillions of hours of video), 
training on all the data is completely impractical. 
Nor is it desirable - most images contain ``typical" 
content and objects that are recognized with 
good accuracy. These models contribute little 
to the final model. But it is the less common 
``interesting" images that are most important 
for training (i.e. images containing objects 
that are mis-classified, or classified with 
low confidence). To benefit from these images, 
it's still important to have accurate label 
information. The images of distance objects 
in isolation are not good for this purpose 
- by definition they contain objects which 
are difficult to label automatically. 
But we can use particular characteristics 
of vehicle video: namely that most of 
the time (vehicle moving forward), objects 
gradually grow and become clearer and 
easier to recognize. We exploit this 
coherence by tracking objects back 
in time and using near, high-reliability 
labels to label more distant objects. 
We demonstrate this process on a
hand-labeled dataset \cite{kitti} which
only has a small fraction of total frames labeled
and also has short length video clips that
can be used for object tracking.
We show that we can extend the labelled 
data using near-to-far tracking strategy, and 
importance sampling can be used to refine
the automatically labeled dataset to improve
its quality. 

To automate far-object tracking, we need 
both single-image object detection and 
between-image tracking. While these two 
modules can be used separately, we designed
a strategy to nicely combine them together.
Specifically we used a Faster-RCNN object 
detector \cite{faster_rcnn} and Kalman 
filtering to track objects. We use predicted 
object positions 
from tracking to augment the Faster-RCNN's 
region proposals, and then use the Faster-RCNN's 
bounding box regression to correct 
the object position estimates. The 
result is that we can track and 
persist object layers much further
into the distance 
where it might be hard for Faster-RCNN
to give accurate object region proposal.

The automatically-generated labels 
are then used to compute the importance of each image.
As shown in \cite{variance_reduction}, 
an optimal model is obtained when images 
are importance sampled using the norm 
of the back-propagated gradient magnitude 
for the image. While computing the full
back-propagated gradients in vehicle video 
systems would be very expensive,
we can actually use the loss function 
as a surrogate for gradient as it is easier
to obtain. We further show in our experiments
that the loss function can be a approximate
of the gradient norm. Importance sampling for training 
data filtering is also described in \cite{is_minibatch}.   
 
\textbf{Contributions}. Starting with a sparsely
labeled video dataset, we combine object tracking
and object detection to generate new 
labeled data, and then use importance 
sampling for data reduction. 
The contributions of this paper are: 
1) We show that near-to-far object 
tracking can produce useful training
data augmentation. 2) Empirically, gradient norm
can be approximated by loss function
and last layer gradient norm in
deep neural networks. 3) Importance 
sampling produces large reductions 
in training data and training time
and modest loss of accuracy. 

\section{Related Work}

\textbf{Semi-Supervised Data Labeling.}
With the large amount of sparsely labeled
image datasets, some work has been done 
in the field of semi-supervised object 
detection and data labeling \cite{tang2016large,
shrivastava2012constrained, lad2014interactively,
choi2013adding}. 
These work try to learn a set of similar 
attributes for image classes 
\cite{shrivastava2012constrained, choi2013adding}
to label new datasets, or to cluster 
similar images \cite{lad2014interactively},
or perform transfer learning to recognize 
similar types of objects \cite{tang2016large}. 
However, these work are typically 
suitable for image dataset where 
they process images individually and 
do not consider the temporal continuity 
of video dataset for semi-supervised learning.
Semi-supervised learning for video dataset 
is also described in \cite{liu2010hierarchical,
prest2012learning,tang2013discriminative}. 
While their performance is good, they assume
labeling the salient object in a video 
and thus do not apply well to multi-object 
detection or tracking case. A large 
body of work has also been done 
in the field of tracking-by-detection \cite{berclaz2011multiple,pirsiavash2011globally,
grabner2008semi,kalal2012tracking,breitenstein2009robust}. 
However, they either assume the possibility of 
negative data being sampled from 
around the object or they do not 
use the special characteristic of 
driving video that objects in 
the near field are easier to be 
detected than objects in the far
field. In addition, these naive 
combinations of tracking and 
detection may introduce additional 
noise in labeling images. Also,
tracking tends to drift away in the long 
run and the related data association 
is also very challenging 
\cite{berclaz2011multiple}. In the 
work of \cite{watch_and_learn}, 
they proposed to use semi-supervised 
learning to label large video 
datasets by tracking multiple 
objects in videos. However, their 
application scenario is not driving 
video dataset and the object they 
detect only include cars. In addition, 
their focus is on short term tracking of 
objects and they do not require short 
tracklets to be associated with each 
other. Therefore, the applicability 
of their method is limited especially 
when there are multiple categories 
of objects in a scene, since 
ignoring the data association part
would be problematic if the goal 
is to label multiple categories 
of objects. In our work, we do 
consider the problem of data association 
and we use object tracker's prediction 
as the region proposal for object 
detector to provide more accurate
bounding box annotation. However, 
similar to the work of \cite{watch_and_learn}, 
we do not perform long-run tracking to
prevent the tracker from drifting away. 
We also use near-to-far labeling to help 
correct the detector's classification results.

\textbf{Importance Sampling.} Importance 
sampling is a well-known 
technique used for reducing the variance 
when estimating properties
of a particular distribution while only 
having samples generated 
from another distribution 
\cite{importance_sampling,rubinstein2011simulation}. 
The work of \cite{importance_sampling1}
studied the problem of 
improving traditional stochastic 
optimization 
with importance sampling, where they 
improved the convergence rate of 
prox-SMD \cite{duchi2009efficient,
duchi2010composite} and prox-SDCA 
\cite{shalev2012proximal} by reducing 
the stochastic variance using 
importance sampling. 
The work of 
\cite{variance_reduction} improves 
over \cite{importance_sampling1} by 
incorporating stochastic gradient 
descent and deep neural networks. 
Also there are some work in using 
importance sampling for minibatch 
SGD \cite{is_minibatch}, where they 
proposed to use importance sampling 
to do data sampling in minibatch 
SGD and this can improve the 
convergence rate of SGD. The idea of hard negative 
example mining is also highly related to
our work. As shown in \cite{shrivastava2016training}
where they presented an approach to perform
efficient object detection training by
training on an optimally sampled bounding boxes
according to their gradient. 

As for self-driving vehicles' vision 
system training, we typically 
do not know the ground truth 
distribution of the data, which 
are the images or video data captured 
by cameras. Thus, importance sampling 
will be very useful in estimating 
the properties of the data from a 
data-driven sampling scheme. 
The work of \cite{DLT} and 
\cite{farah2014computationally} proposed 
to use importance sampling for 
visual tracking, but their focus was not 
on reducing the training data amount
and creating labeled data using visual 
tracking. In our work, we use importance 
sampling to obtain an optimal set of 
data so that our training efficiency 
is high as we train
on the most informative data. The 
information that each image carries 
is characterized 
by their detection loss, which is 
reasonably suitable in our case as 
images with high loss are usually
images that are difficult for 
the current detector.

\section{Methods}

Our approach for creating labeled data 
and performing data reduction by importance 
sampling can be divided into two parts. 
First of all, based on the sparsely labeled 
image frames, we initialize object tracker 
by incorporating Kalman-filter algorithm 
\cite{welch1995introduction} and use the 
tracker to predict bounding box of objects 
in the previous (since we predict back in time) frame. We then use the prediction 
as region proposal input and send this to 
object detection module. Based on the 
region proposal received, the object 
detection module trained on sparsely 
labeled data will do a bounding box 
regression to get the final bounding 
box and detection loss. The object 
tracker further matches new detections 
to existing trackers or create new 
trackers if the new detection cannot 
match any of the existing trackers. 
The near-to-far labeling module will 
double check object detection results
within each tracker to use the 
classification results of objects in 
the near field which are more accurate 
to check the results of objects in the 
far field. The bounding box produced by 
the object detection module are used as
labels for those unlabeled video frames. 
The detection loss will be used as the
sampling weights for importance sampling.
Secondly, based on the detection loss
recorded in the first part, the importance
sampler will sample an optimal subset of
labeled images and these selected labeled 
images will be used as the training data 
to train a new object detector.

The system architecture is shown 
in figure~\ref{fig1}. Here, 
we first describe the framework of 
semi-supervised data labeling followed by
the data reduction using the 
importance sampling framework.

\begin{algorithm}[t]
 \textbf{Description:}Update object bounding box 
 label for a single track-back-in-time step.
 
 \KwIn{$\mathcal{T}$: the list of active trackers;
        $l_{\epsilon}$: threshold 
        of detection loss, save a detection if the 
        detection loss is larger than this threshold; 
        $\mathcal{K}$: Kalman filter; $\mathcal{D}$: 
        object detector; $n$: limit of steps to retain
        a tracker;\\}
 \KwResult{$\mathcal{R}$ (collections of detections 
 to be saved, a detection is a bounding box containing
 state information $(x,y,s,r, label)$ where $label$
 is the class of the object)\\ }
 \textbf{Initialize: }{ $\mathcal{R}=\varnothing$,
 $\mathcal{R}'=\varnothing$, $\mathcal{P}_m=\varnothing$, 
 $\mathcal{P}_m$ is the matched pairs of detection and tracker.\\}
\For{$\forall$ $T$ $\in$ $\mathcal{T}$}
    { /* Tracking back in time */\\
    $s$ = predict\_state($T$, $\mathcal{K}$); \\
    /* Use $s$ (prediction of RoI) as region proposal */ \\
    $d$ = get\_new\_detection($s$, $\mathcal{D}$); \% Refer to section 3.1\\
    $\mathcal{R}'$ = $\mathcal{R}'\bigcup d$;}
\For{$\forall$ $d, T$ $\in$ $\mathcal{R}', \mathcal{T}$}
    {Match $d$, $T$\\ 
    \If{$d,T$ \text{can match}}
    {$\mathcal{P}_m = \mathcal{P}_m\bigcup (d,T)$;}
    }
$\mathcal{R}'_{um}$= $\{d; d\in \mathcal{R}' 
\quad\&\quad d\notin \mathcal{P}_m\}$ (get unmatched $d$); \\
$\mathcal{T}_{um}$= $\{T; T\in \mathcal{T}\quad
\&\quad T\notin \mathcal{P}_m\}$ (get unmatched $T$);\\
\For{$\forall$ $(d, T)\in\mathcal{P}_m$}
    {Update tracker $T$ using $d$ using Kalman Filter; \\ 
    Using historical records in $trk$ to check 
    the accuracy of this new detection $d$; \\
    $\mathcal{R} = \mathcal{R}\bigcup d$;}
\For{$\forall$ {\rm unmatched} $d\in\mathcal{R}'_{um}$}
    {$T$ = init\_new\_tracker($d$);\\
    $\mathcal{R}$=$\mathcal{R}\bigcup d$; }
\For{$\forall$ {\rm unmatched} $T\in\mathcal{T}_{um}$}
    {\If{$T$ {\rm has not been updated for more than $n$ times}}
        {Remove $T$ from $\mathcal{T}$}
        }
\For{$\forall$ {\rm d} $\in\mathcal{R}$}
    {\If{$d$ \rm has loss $> l_{\epsilon}$} 
     {mark this $d$ to be saved }
    }
return $\mathcal{R}$
 \caption{Object Tracking and Labeling Algorithm}
 \label{alg1}
\end{algorithm}

\begin{algorithm}[t]
\textbf{Description:} Match detections with trackers. 

\KwIn{$\mathcal{R}'$: object detection bounding boxes, $d=[x_1, y_1, x_2, y_2]$; $\mathcal{T}$: trackers;\\}

\KwResult{Matched detection and tracker pairs $\mathcal{P}_m$; Unmatched detections $\mathcal{R}'_{um}$; Unmatched Trackers $\mathcal{T}_{um}$;}
\If{len($\mathcal{T}$) == 0}{
    $\mathcal{P}_m=\varnothing$;
    $\mathcal{R}'_{um}=\mathcal{R}'$;
    $\mathcal{T}_{um}=\varnothing$;\\
    return $\mathcal{P}_m, \mathcal{R}'_{um},\mathcal{T}_{um}$.
}
$M$ = An all-zeros matrix of size $[len(\mathcal{R}'), len(\mathcal{T})]$.\\
\For{$i$ \text{from 1 to len}($\mathcal{R}'$)}{
\For{$j$ \text{from 1 to len}($\mathcal{T}$)}{
$M[i,j]$ = IntersectionOverUnion($\mathcal{R}'[i], \mathcal{T}[j]$)
}
}
$M2$ = linear\_assignment($M[i,j]$);\\
\For{$i$ \text{from 1 to len}($\mathcal{R}'$)}{
    \If{$i\notin M2[:,0]$}{
        $\mathcal{R}_{um}' = \mathcal{R}_{um}'\bigcup\mathcal{R}'[i]$
    }
}
\For{$j$ \text{from 1 to len}($\mathcal{T}$)}{
    \If{$j\notin M2[:,1]$}{
        $\mathcal{T}_{um} = \mathcal{T}_{um}\bigcup\mathcal{T}[j]$
    }
}
\For{$i,j \in M2$}{
    $\mathcal{P}_m = \mathcal{P}_m\bigcup(\mathcal{R}'[i], \mathcal{T}[j])$.
}
return $\mathcal{P}_m, \mathcal{R}'_{um},\mathcal{T}_{um}$.
\caption{Match Detections with Trackers}
\label{alg2}
\end{algorithm}

\subsection{Semi-supervised Data Labeling}
\textbf{Object Tracking.} Starting with a few sparsely 
annotated video frames, we first trained an object detection
network using Faster-RCNN \cite{faster_rcnn}. By using 
Kalman filter \cite{welch1995introduction},
we initialize object trackers with the ground truth 
labeled frames. The specific object tracking framework
we use is similar to that of \cite{Bewley2016_sort}, 
where the state of the tracker includes 7 parameters, 
namely, the center position of the bounding box $x,y$, 
the scale $s$ (the area of the bounding box) and aspect ratio $r$ of the bounding box ( the ratio of the width over the height of the bounding box), and the
rate of change of the center position $v_x, v_y$ and 
scale $v_s$ of the bounding box. We follow the assumption in
\cite{Bewley2016_sort} that the aspect ratio of the 
bounding box does not change over time. The measurement 
is just the first four parameters of the state vector. 
\begin{equation*}
\text{state} = [x, y, s, r, v_x, v_y, v_s] \\
\end{equation*}
\begin{equation*}
\text{measurement} = [x, y, s, r]
\end{equation*}
We always use ground truth labeled bounding box to 
initialize object trackers, and the tracking is done 
from the near field to the far field, which means the 
video is played in the opposite direction as it was 
collected, so that at the very beginning, the camera is
close to the labeled objects and at the very end the 
camera is far away from the the object. Therefore, it is
reasonably to believe that the classification and detection 
results for objects in the near field are more reliable while
there is more noise in the detection results for object in the
far field. 

\textbf{Prediction as Region Proposal. } After the trackers are
initialized with ground truth bounding box, based on the principle
of a Kalman filter, predictions of bounding boxes of the objects 
being tracked will be calculated. These predictions will be used
as a hint for the object detection network to produce new bounding 
boxes in the next frame. The network we used for object detection 
is Faster-RCNN \cite{faster_rcnn}, which is composed of a region 
proposal network (RPN) and object detection network Fast-RCNN 
\cite{fast_rcnn}. Usually the RPN will be used as the region proposer,
however, as we already have the prior information of where the object
might be, we can directly use this information to help the object 
detection network avoid uncertainty in
region proposal. This part corresponds to the \texttt{get\_new\_detection}
method in algorithm~\ref{alg1}.

\textbf{Matching Tracker with Detections.} Given the predictions sent
by the object tracker, the object detection network will produces a 
set of candidate bounding boxes in the next frame and the object tracker
will try to match the existing trackers and the new detections using linear
assignment. We also use intersection over Union (IoU) to filter out 
detection-tracker pairs that do not have IoU values higher than a pre-defined
threshold. After finishing detection-tracker matching, the state of valid
trackers will be updated, and trackers that remain inactive (not being updated)
for a certain steps will be removed from the trackers list. Now we finished one step
of object tracking and labeling. The bounding boxes produced by object detection
network will be used as labels for those unlabeled video frames. 
The more detailed algorithm description for one step of tracking
and labeling is shown in algorithm~\ref{alg1}. Matching trackers with
detections is further described in algorithm~\ref{alg2}.

The \textit{tracker} is a class containing state of the current 
object being tracked and methods for updating object's state given
ground truth state of the object.  
A detailed implementation of the tracker class
can be found in~\cite{kalman}.

\textbf{Near-to-Far Labeling}. Another key ingredient of our approach 
is the near-to-far labeling scheme. Consider the case that we are tracking 
an object from far to near field. When the image is far away from our 
current location, the object could be very small or blurred in the image, 
which makes it very difficult to be correctly classified. As the object 
approaches the vehicle, the detection network has a higher confidence to 
correctly classify this object. As we trust object detection results 
in the near field, if object detection results of the same object being 
tracked in the far field differ from that in the near field, we can use
the detection results in the near field to correct that. To do this, we 
restrict object tracker's initialization only to ground truth bounding 
boxes so as to avoid the additional noise introduced by imperfect object
detection network. In case the classification of objects in the far field
diverges, we use the detection result of the same tracker in the near 
field to correct that. Examples of near-to-far labeling are shown in 
figure~\ref{implicit}.


\subsection{Sampling an Optimal Subset of Images}
Inspired by the idea of importance sampling \cite{variance_reduction}, 
we can select an optimal subset of the data by sampling the data 
according to importance sampling probability distribution so that 
the variance of the sampled data is minimized under an expected 
size of sampled data. Here, the sampling distribution is proportional 
to the object detection loss of each image. Images with
higher loss obtain more importance as they provide more useful
information for accurate object detection. 

In our case, we are interested in estimating the expectation of $f(x)$ based on 
a distribution $p(x)$, where $f(x)$ is the detection loss of each image, $p(x)$
denotes the image distribution and $x$ denotes a particular image with an object 
detection loss. The problem is expressed by the following equation,
\begin{equation}
\int{p(x)f(x)\text{d}x} = \mathbb{E}_{p(x)}[f(x)] \approx \frac{1}{N} \sum_{n=1}^Nf(x_n),
\end{equation}
where $x_n \sim p(x)$. 
However, usually we do not know the ground truth distribution 
of the data $p(x)$, so we rely on a sampling proposal $q(x)$ to  to 
unbiasedly estimate this expectation, with the requirement that 
$q(x) > 0$ whenever $p(x) > 0$. This is commonly known as importance 
sampling: 
\begin{equation}
\int{p(x)f(x)}dx = \mathbb{E}_{p(x)}[f(x)] = \mathbb{E}_{q(x)}[\frac{p(x)}{q(x)}f(x)].
\end{equation}
It has been proved in \cite{variance_reduction} that the variance 
of this estimation can be minimized when we have,
\begin{equation}
q^{*}(x) \propto p(x)|f(x)|.
\end{equation} 
Defining $\tilde{q^{*}}(x_i)$ as the unnormalized optimal probability
weight of image $x_i$, it is obvious that images with a larger detection loss
 should have a larger weight. Although we do not know $p(x)$, 
we have access to a dataset $\mathcal{D} = \{x_n\}_{n=1}^{N}$ sampled from $p(x)$. 
Therefore, we can obtain $q^{*}(x)$ by associating the unnormalized probability weight 
$\tilde{q^{*}}(x_n) = |f(x_n)|$ to every $x_n \in \mathcal{D}$, 
and to sample from $q^{*}(x)$ we just need to normalize these weights:
\begin{equation}
q^{*}(x_n) = \frac{\tilde{q^{*}}(x_n)}{\sum_{i=1}^N \tilde{q^{*}}(x_i)} 
= \frac{|f(x_n)|}{\sum_{i=1}^N|f(x_i)|}
\end{equation}
where $f(x_i)$ is the loss of input $x_i$. To 
reduce the total number of data instances used for estimating 
$\mathbb{E}_{p(x)}[f(x)]$, we draw $M$ samples from the whole $N$ data 
instances ($M << N$) based on a multinomial distribution where 
$(q^{*}(x_1),...,q^{*}(x_N))$ are the parameters of this multinomial 
distribution. Based on the discussion 
above, we obtained an estimation of  $\mathbb{E}_{p(x)}[f(x)]$ which 
has least variance compared to all cases where we draw $M$ samples 
from the entire $N$ data set. We further provide some prove in the appendix.

\begin{figure*}[t]
 \centering
\includegraphics[width=0.2\linewidth,height=0.2\linewidth]{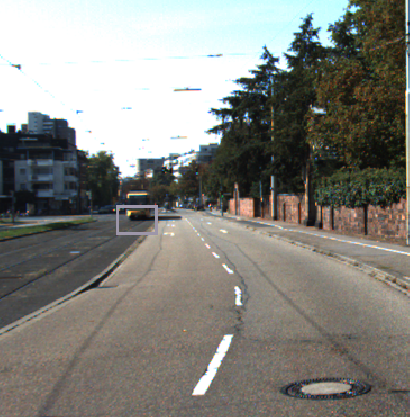}
\includegraphics[width=0.2\linewidth,height=0.2\linewidth]{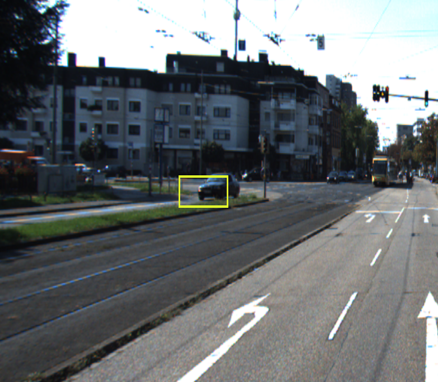}
\includegraphics[width=0.2\linewidth,height=0.2\linewidth]{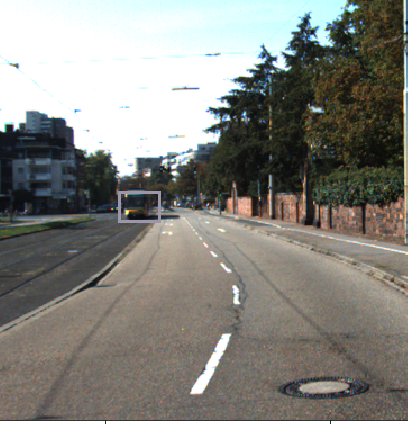}
\includegraphics[width=0.2\linewidth,height=0.2\linewidth]{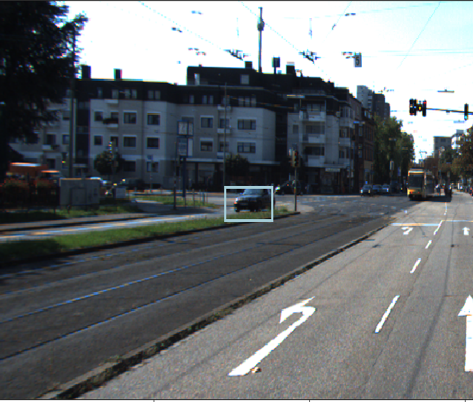}
\caption{Bounding box generated by using strategies mentioned in experiment 1 
(left 1 and 2) and experiment 2(right 1 and 2)}
\label{fig2}
\end{figure*}

\begin{table*}\centering
\ra{0.90}
\begin{tabular}{@{}p{3cm}ccccccccccccc@{}}\toprule
& \multicolumn{3}{c}{Pedestrian} & \phantom{abc} & \multicolumn{3}{c}{Car} & \phantom{abc} & 
\multicolumn{3}{c}{Cyclist} & mAP\\ \cmidrule{2-4} \cmidrule{6-8} \cmidrule{10-12}
& Easy & Medium & Hard && Easy & Medium & Hard && Easy & Medium & Hard &\\\hline
Ground Truth (GT) & 
80.6 & 68.8 & 61.0 && 
94.1 & 78.8 & 69.3 && 
88.1 & 78.8 & 73.6 & 77.0 \\
New Labeled (NL) \& GT &
69.2 & 58.4 & 50.8 &&
83.4 & 63.2 & 53.1 &&
68.3 & 56.6 & 52.9 & 61.8 \\
Sampled NL \& GT&
\textbf{71.3} & \textbf{62.7} & \textbf{54.1} &&
75.8 & 61.5 & 51.6 &&
\textbf{77.5} & \textbf{66.0} & \textbf{61.3} & \textbf{64.6} \\
Only NL & 
69.8 & 60.8 & 52.1 &&
80.4 & 60.9 & 50.3 &&
70.4 & 57.8 & 54.0 & 61.8 \\
\bottomrule
\end{tabular}
\caption{Object detection average precision (\%) on KITTI dataset using 
different models. Ground Truth: results of model trained on ground
truth labeled data (comes from KITTI). New labeled and ground truth: 
model trained on both new labeled data and ground truth data, corresponding
to experiment 1. Sampled
NL and GT: model trained on importance sampled new labeled data and ground
truth data, corresponding to experiment 2. Only NL: model trained
only on new labeled data, corresponding to experiment 3. }
\label{mAP}
\end{table*}

\begin{figure*}[t]
\centering
\includegraphics[width=\linewidth,height=0.24\linewidth]{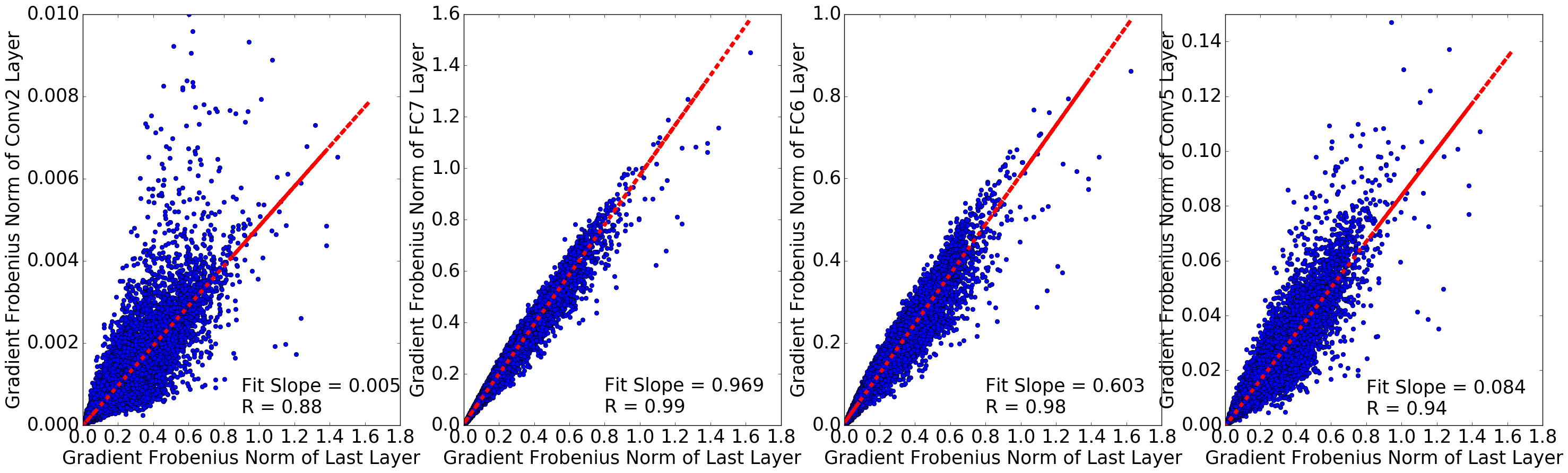}
\caption{Plot of gradient Frobenius norm of last layer 
in VGG 16 versus 
the gradient Frobenius norm of fully-connected layer 7 (FC7), 
fully-connected layer 6(FC6), Convolutional Layer 5 (Conv5) 
and Convolutional Layer 1 (Conv1).}
\label{fig:linear}
\end{figure*}

\begin{figure*}[t]
\centering
	\includegraphics[width=0.3\linewidth, 
    height=0.18\linewidth]{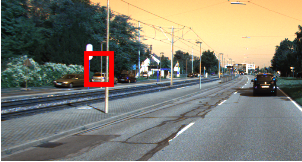}
    \includegraphics[width=0.3\linewidth, 
    height=0.18\linewidth]{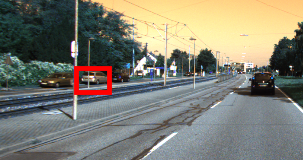}
    \includegraphics[width=0.3\linewidth, 
    height=0.18\linewidth]{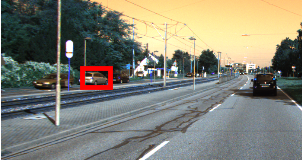}
  	\includegraphics[width=0.3\linewidth,
    height=0.18\linewidth]{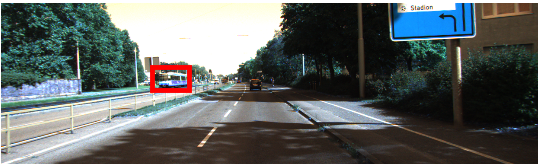}
    \includegraphics[width=0.3\linewidth,
    height=0.18\linewidth]{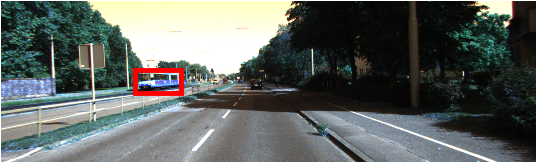}
    \includegraphics[width=0.3\linewidth,
    height=0.18\linewidth]{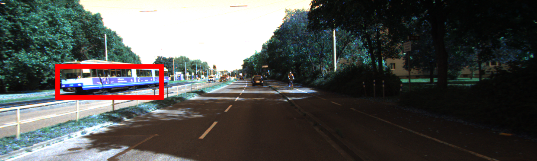}
 	\includegraphics[width=0.3\linewidth,
    height=0.18\linewidth]{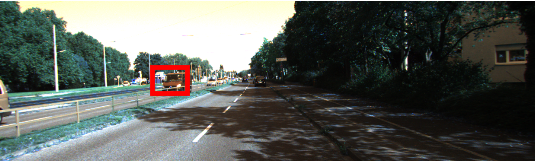}
    \includegraphics[width=0.3\linewidth,
    height=0.18\linewidth]{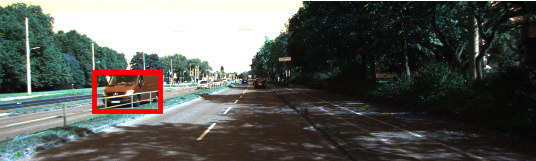}
    \includegraphics[width=0.3\linewidth,
    height=0.18\linewidth]{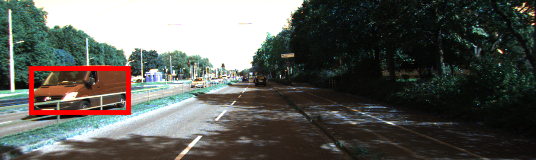}
     	\includegraphics[width=0.3\linewidth,
    height=0.18\linewidth]{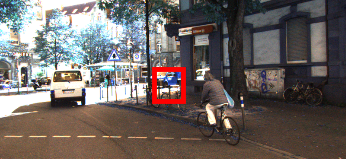}
    \includegraphics[width=0.3\linewidth,
    height=0.18\linewidth]{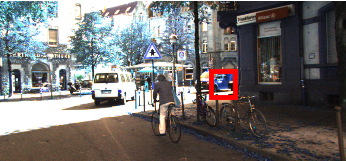}
    \includegraphics[width=0.3\linewidth,
    height=0.18\linewidth]{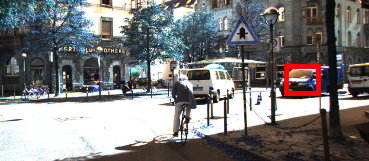}
    \caption{Examples of near-to-far labeling. These images are from the KITTI Benchmark data set \cite{kitti}. The labeling results are obtained from pre-trained Faster-RCNN model. The bounding box shows the detected objects being tracked. Near field object detection results are used to check the accuracy of the detection results of objects in the far field. The first image labeling results  from left to right: motorbike, car, car (ground truth). As the vehicle approaches the object, it becomes clearer and no longer hidden by the pole. The second image labeling results from left to right: bus, car, train (ground truth). The object looks like a bus in the far field, but is classified as train when it's in the near field considering it's on railroad. The third image labeling results from left to right: train, car, car (ground truth). The object looks like a train in the far field, but is classified as car when it's in the near field. Fourth image labeling results from left to right: bus, car, car (ground truth). At first sight, the car is blurred and hidden by other objects, then it became more clearer that this is a car. 
    }
    \label{implicit}
\end{figure*}

\subsection{Measuring Variance Reduction Efficiency}\label{ssec:num3}
Once we get the sampling distribution 
$q^{*}(x_i)$, we then perform the 
importance sampling. Images with a higher 
detection loss will get higher 
likelihood to be sampled. We, further, 
measure how efficient that we estimate 
the detection loss
distribution. Since the goal of using 
importance sampling approach 
here is to reduce the variance while 
estimating properties of the 
data from a subset of the data. 

To show that the expectation of loss 
estimated from the sampled images have close 
variance with loss variance estimated 
from all images, we computed 
a relative variance value. This value
is the ratio of whole data set detection 
loss variance over sampled images' 
detection loss variance. 

Suppose the data set is $\mathcal{D} 
= \{x_n\}_{n=1}^N$, and we can get 
detection loss $g(x_i)$ given individual 
input $x_i$. In order to calculate the 
relative variance more easily, we will 
first normalize $g(x)$. Then, we define 
the sampling probability of image $x_i$ 
when we expect to sample M out of N images
($M < N$) as, 
\begin{equation} \label{eq:qi}
q(x_i) = \min\bigg[1, \frac{
M|g(x_i)|}{\sum_{i=1}^N|g(x_i)|}\bigg]
\end{equation}
taking the minimum compared with 1 
is to ensure that the probability of
sampling image $x_i$ can not be larger
than 1, which happens when 
$\frac{
M|g(x_i)|}{\sum_{i=1}^N|g(x_i)|}$ is 
saturated. Note that,
when the sampling probability is 1, 
we should sample this image.
With the scaled sampling weight 
$\frac{
M|g(x_i)|}{\sum_{i=1}^N|g(x_i)|}$, 
we change $
M$ so that we  
can get different numbers of images 
out of the entire image date. 
Typically, choosing a $M$ such that 
the sample gradient norm
variance is close to whole data 
gradient norm variance. 
Since the data are in the discrete 
space, the relative variance is 
defined as,

\begin{equation} 
\begin{split}
{\rm R} 
& = \frac{\sum_{i=1}^N|g(x_i)|^2}{\sum_{i=1}^N|g(x_i)|^2/q(x_i)}.\\
\end{split}
\label{ratio}\end{equation}

\begin{figure}
\centering
\includegraphics[width=0.7\linewidth]{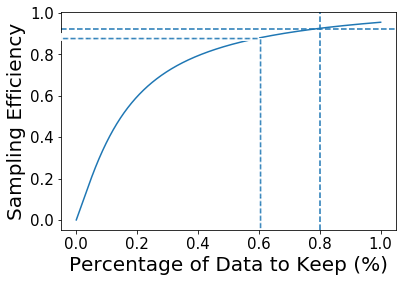}
\caption{Relative Variance Evaluation Results}
\label{eff}
\end{figure}

\section{Experiments}

Our framework has several major contributions.
First of all, we proposed to use
object tracker's prediction as the region 
proposal input for the object detection
network to detect objects. Secondly, we 
proposed to use near-to-far labeling to
help correct labels that may not be correct. 
Thirdly, we use importance sampling 
to select an optimal subset of images to 
remove images with less reliable labels
and obtain a smaller but more informative 
set of data. We designed several comparative
experiments to show the impact of our contribution.

\subsection{Comparative Experiments}
\textbf{Datasets}. To show that our algorithm 
is able to scale to a relatively large video
dataset, we choose the KITTI benchmark dataset 
\cite{kitti} which contains hundreds of 
autonomous driving video clips, and each of 
the video clips lasts about 10 to 30 seconds. 
The data set is fairly rich as it contains 
high-resolution color and grayscale video frames 
captured in many kinds of driving environments: 
city, residential, road, campus, person, etc. 
The KITTI dataset also contains a set of 
sparsely labeled image frames for object 
detection purposes. The number of images 
with ground truth bounding box labeling we 
used in our experiment is 7481, while the 
total number of images is around 40000. 
Categories of objects being labeled include 
cars, pedestrians, vans, trams, cyclist, 
truck, person sitting, and so on. For 
simplicity, we choose 3 categories from 
them to detect, which include cars, 
pedestrians, and cyclist. We manually and 
randomly divide 
the dataset into the training, validation 
and test data set. The training dataset 
contains 4206 images, the validation dataset
contains 1404 images, and the test data 
set contains 1871 images. 

\textbf{Experiment 0}. We first trained 
a basic object detection network based 
on the ground truth labeled data using 
the Faster-RCNN \cite{faster_rcnn} object 
detection network. As for details of 
training, we used pre-trained Faster-RCNN
model with VGG16 network \cite{vgg16} 
trained on PASCAL VOC 2007 dataset 
\cite{everingham2010pascal}, and then 
finetuned with KITTI dataset. The number of 
training iterations is 300k with the initial
learning rate of 0.01 and decay every 30k 
iterations. 

\textbf{Experiment 1}. 
The first experiment is our labeling by tracking approach 
using semi-supervised learning.
In this experiment, we use the ground truth 
labeled bounding boxes to initialize object
trackers. Since images in the KITTI dataset 
are sparsely labeled with unlabeled images 
between labeled images in the original 
video sequence, we use the labeled data 
as a guidance to label images 
without ground truth labeling. It is useful 
to notice that, in this case, the object 
detection network does not use RPN to generate
region proposals. Instead, it takes
the object tracker's prediction of bounding 
box in the next frame as region proposal 
and then perform bounding box regression
to generate optimal bounding box for the 
object being tracked. In other words,
only ground truth labeled images can 
be used to initialize object tracker, 
which is based on our 
assumption that objects in the near 
field provide more accurate information 
and we only predict bounding boxes based
on reliable information instead of
relying on some random detection. We used
both ground truth data from KITTI combined
with new labeled data to train the object 
detector. The training setting is the same
as in experiment 0.

\textbf{Experiment 2}. In this experiment, 
we adopt the approach we take in experiment 1 
and we further combine it with importance 
sampling. As images labeled using the 
approach in experiment 1 may still contain 
redundant information such as images
that are already easy for the network to 
process, so we use importance 
sampling to select an optimal set of images 
that are more informative. 
We choose to sample 60\% of the data (
which consists of both ground truth and new
labeled data) in experiment 1 using the importance sampling 
method mentioned in previous section. 
As shown in figure~\ref{eff}, 60\% of 
data corresponds to around 0.90 sampling efficiency, 
which is reasonably high. The training setting is 
also the same as in experiment 0.

\textbf{Experiment 3}. We further remove
the ground truth data which comes from
KITTI and only used newly labeled data 
using to train an object detector with 
the same training setting as in experiment
0. 

\textbf{Evaluation of Accuracy}. We trained 
Faster-RCNN object detection networks using 
data mentioned in experiment 1, 2, and 3 
respectively, all using the same training 
configurations as we did in experiment 0. 
We evaluate the performance of models in
experiment 0,1,2,and 3 by testing the models 
on a held out test dataset of 1871 images. 
The average precision is evaluated on the 3
categories of objects mentioned before. 
\subsection{Results and Analysis}

\begin{figure}[h]
    \centering
    \includegraphics[width=0.8\linewidth]{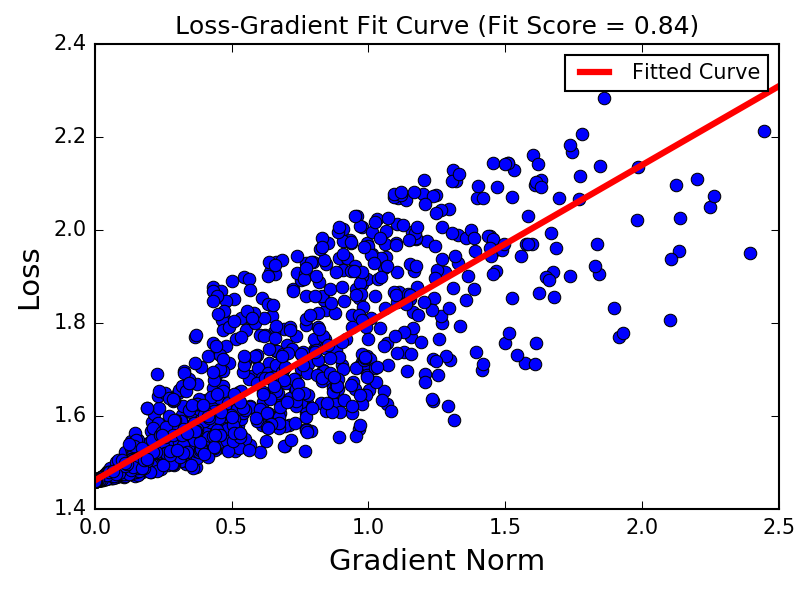}
    \caption{Plot of gradient Frobenius norm of entire network
    weight (not include bias term) VS the loss of individual data point.}
    \label{fig:loss-gradient}
\end{figure}

\textbf{Loss as a Approximation for Gradient}
First, we show our finding that the gradient
of the network we used has some linear 
correlation between different layers as shown
in figure~\ref{fig:linear}. Therefore,
we can use last layer gradient (as it is easier
to obtain) as a approximation of total gradients.
On the other hand, we also show in
figure~\ref{fig:loss-gradient} that loss
can be used as a approximation for the total gradient
norm. Therefore, we can also use loss to 
approximate gradient and use it as sampling
weight for different object bounding box labels.

\textbf{Qualitative Results for Bounding 
Box Generation}. As mentioned in experiment 
description, we use two different 
strategies to generate bounding boxes 
using Faster-RCNN. The first strategy 
uses region proposal network 
to generate bounding boxes, and the 
second strategy uses object tracker's 
prediction as region proposals. We show 
some qualitative results of bounding boxes 
generated by the two methods in figure 
~\ref{fig2}.

\textbf{Quantitative Results for Object 
Detection}. The accuracy of models 
trained on experiment 0,1,2,and 3 are 
evaluated on a test data set of 1871 images. 
The average precision (AP) on 3 categories
of objects and the mAPs are reported in 
table~\ref{mAP}. The results show the
average precision for different
categories of objects with different
degrees of difficulty. With the 
ground truth data, the 
model shows the best performance,
which is not a surprise since
labels generated by tracking 
may introduce noise that harms
the performance of the detector.
However, after filtering the 
data by importance sampling,
we can obtain better detection
accuracy using the same training
setting, which means importance 
sampling has helped to reduce 
data volume and makes it easier
to train a model to convergence.

\textbf{Relative Variance Results}
We use the relative variance mentioned in 
section \ref{ssec:num3} to measure 
how good we estimate the image detection 
loss distribution. The result is shown here
~\ref{eff}. From the plot, we can see that 
by scaling the importance sampling weight 
as mentioned in \ref{eq:qi}, we are able 
to keep high sampling efficiency (0.90)
with 60 \% of the original labeled data 
being sampled. This curve will be useful 
for determining how much data to sample given
the desired sampling efficiency.

\section{Conclusion}
We proposed a framework of automatically 
generating object bounding box labels
for large volume driving video dataset 
with sparse labels. Our work generates
object bounding boxes on the new labeled data 
by employing a near-to-far labeling strategy,
a combination of object tracker's 
prediction and object detection network 
and the importance
sampling scheme. Our experiments show
that with our semi-supervised learning framework,
we are able to annotate driving video 
dataset with bounding box labels and 
improve the accuracy of
object detection with the new labeled data
using importance sampling. 


{\small
\bibliographystyle{ieee}
\bibliography{egbib}
}
\clearpage
\begin{appendices}
\section{Introduction}
We provide proof of the importance sampling framework and their optimality in
this supplementary material. We also provide detailed explanations for the 
measurement of relative variance and the meaning of relative variance. 

\section{Importance Sampling Framework Proof}

The importance sampling algorithm is used for data reduction. It is also used 
for the selection of an optimal subset of data from the original labeled 
dataset with minimal variance. In the paper, 
we stated that by using a 
reference proposal distribution $q^{*}(x) \propto p(x)|f(x)|$ we can get 
an estimation of the expectation of $f(x)$ with the least variance. 
We now provide the proof.

In importance sampling, the expectation of $f(x)$ is
estimated by using $\mathbb{E}_{p(x)}[f(x)] = \mathbb{E}_{q(x)}[f(x)p(x)/q(x)]$. We require that $q(x) >0$
whenever $f(x)p(x)\neq 0$. It is thus easily to verify that this estimation is unbiased. Suppose that 
$\mathbb{E}_{p(x)}[f(x)]$ is defined on 
$x\in A$ while $\mathbb{E}_{q(x)}[f(x)p(x)/q(x)]$ is 
defined on $x\in B$. We have $A = \{x| p(x) > 0\}$ and $
B = \{x | q(x) > 0 \}$. So that we have for $x\in A\cap B^c$
, $f(x) = 0$ and for $x \in A^c\cap B$, $p(x) = 0$. That is
to say, for $x \in A\cap B^c$ and $x \in A^c\cap B$, we have
$f(x)p(x) = 0$. So the expectation of $f(x)$ can be written 
as, 
\begin{equation}
\begin{split}
\mathbb{E}_{q(x)}[\frac{p(x)f(x)}{q(x)}] 
= &\int_{B}\frac{f(x)p(x)}{q(x)}q(x)\text{d}x \\
= & \int_{A}f(x)p(x)\text{d}x +
\int_{B\cap A^c}f(x)p(x)\text{d}x\\
& -\int_{A\cap B^c}f(x)p(x)\text{d}x \\
= &\int_{A} f(x)p(x)\text{d}x \\
= & \mathbb{E}_{p(x)}[f(x)]
\end{split}
\end{equation}
Then we prove that when sampling
distribution $q(x) \propto p(x)|f(x)|$,
we can obtain the minimal variance in
the estimation of the expectation. Let 
$\mathbb{E}_{p(x)}[f(x)] = \mu$, and let,
\begin{equation}
{\mu}_q = \frac{1}{n}\sum_{i=1}^n \frac{f(x_i)p(x_i)}{q(x_i)}
\end{equation}
given samples $x_i$ are sampled from $q(x)$. Then the 
variance of $\mu_q$ is,
\begin{equation}
\begin{split}
\mathrm{Var}({\mu}_q)& = \frac{1}{n}\mathrm{Var}\bigg(\frac{f(x_0)p(x_0)}{q(x_0)}\bigg) \\
& = \frac{1}{n}\bigg(\int (f(x)p(x))^2/q(x) \text{d}x - \mu^2\bigg)\\
\end{split}
\end{equation}
By choosing $q^{*}(x) = |f(x)|p(x)/\mathbb{E}_p(|f(x)|)$,
and let $q(x)$ be any density function that is positive given
$f(x)p(x)\neq 0$. We have,
\begin{equation}
\begin{split}
\mathrm{\mathrm{Var}}({\mu}^{*}_q) & = \frac{1}{n}\bigg(\int\frac{(f(x)p(x))^2}{q^{*}(x)}\text{d}x - \mu^2\bigg) \\
& = \frac{1}{n}\bigg(\int\frac{(f(x)p(x))^2}{|f(x)|p(x)/\mathbb{E}_p(|f(x)|)}\text{d}x - \mu^2\bigg) \\
& = \frac{1}{n}\bigg(\mathbb{E}_p(|f(x)|)^2 - \mu^2\bigg) \\
& = \frac{1}{n}\bigg(\mathbb{E}_q(|f(x)|p(x)/q(x))^2 - \mu^2\bigg) \\
& \leq \frac{1}{n}\bigg(\mathbb{E}_q(f(x)^2p(x)^2/q(x)^2) - \mu^2\bigg) \\
& = \mathrm{\mathrm{Var}}({\mu}_q)
\end{split}
\end{equation}
The last inequality is the Cauchy-Schwarz inequality. Therefore, we show that
by choosing sampling distribution $q(x) \propto p(x)|f(x)|$ and sampling data 
according to the normalized $q_{normalized}(x_i) = q(x_i)/\sum{q(x_i)}$, we can obtain the minimal variance estimation. In the case where $p(x)$ is not known directly,
but we have a dataset sampled from $p(x)$, we can use $q(x_i) = |f(x_i)|/
\sum_i{|f(x_i)|}$ as the sampling weight.

\section{Measuring the Efficiency of Sampling}
We define the efficiency as the ratio between the original data variance 
and the sampled data variance. To make things simpler,
suppose we want to estimate the expectation of $f(x)$, we first
normalize $f(x)$ and obtain $g(x) = (f(x) - \overline{f(x)})/\sigma{[f(x)]}$, 
where $\overline{f(x)}$ and $\sigma{[f(x)]}$ are the mean and standard deviation of $f(x)$. 
Now we use importance sampling to estimate the expectation of $g(x)$ under $p(x)$ 
by using proposal distribution $q(x)$. We sample
$M$ images out of a total $N$ images, the probability of a particular image $x_i$
being sampled is,
\begin{equation}
s(x_i) = \min\bigg[{1, \frac{M|g(x_i)|}{\sum_i^{N}{|g(x_i)|}}}\bigg]
\end{equation}
As mentioned in the paper, we take the minimum compared with 1 to ensure that the 
probability is always no more than 1. Obviously, $\sum_{i=1}^N s(x_i)>1$ since $s(x_i)$ 
describes the probability of a particular image $x_i$ being selected. We further define
$q(x_i) = s(x_i)/N$ which has an upper bound of $1/N$. Therefore, it is easy to see that 
$\sum_{i=1}^Nq(x_i) \leq 1$.  To get $M$ images, we select images according
to their sampling probabilities $s(x_i)$. The expectation of
$g(x)$ based on the sampled images is,
\begin{equation}
\mathbb{E}_{q(x)}[g(x)p(x)/q(x)] =\frac{1}{N} \sum_{i=1}^Ng(x_i)p(x_i)/q(x_i)
\end{equation}
where $x_i \sim q(x)$. On the other hand, if we sample the entire
dataset and get $N$ images, then $s(x_i) = 1$ and $q(x_i) = 1/N$, the expectation will be,
\begin{equation}
\mathbb{E}_{q(x)}[g(x)p(x)/q(x)] = \frac{1}{N}\sum_{i=1}^Ng(x_i)p(x_i)*N
\end{equation}
which is just $\mathbb{E}_{p(x)}[g(x)]$. It is no harm to assume $p(x)$ is a uniform 
distribution since we consider it to be unknown. In the case where we sample
the entire dataset, $s(x_i)= 1$, $p(x_i) = 1/N$, and $\sum_{i=1}^Ng(x_i)=0$, then
the variance of $g(x)$ by sampling the entire dataset is,
\begin{equation}
\begin{split}
\mathrm{Var}_{q}\bigg[\frac{g(x)p(x)}{q(x)}\bigg] = &  \mathbb{E}_{q}\bigg[\big(\frac{g(x)p(x)}{q(x)}\big)^2\bigg] -\bigg(\mathbb{E}_{q}\bigg[\frac{g(x)p(x)}{q(x)}\bigg]\bigg)^2 \\
= & \sum_{i=1}^N\bigg[\frac{g(x_i)p(x_i)}{s(x_i)/N}\bigg]^2\frac{s(x_i)}{N} \\
& - \bigg(\sum_{i=1}^N\bigg[\frac{g(x_i)p(x_i)}{s(x_i)/N}\bigg]\frac{s(x_i)}{N}\bigg)^2 \\
= & \sum_{i=1}^N\bigg[\frac{g(x_i)/N}{1/N}\bigg]^2\frac{1}{N} \\
& - \bigg(\sum_{i=1}^Ng(x_i)/N\bigg)^2 \\
= & \sum_{i=1}^Ng^2(x_i)/N
\end{split}\label{var1}
\end{equation}
In the case where we sample $M$ images out of $N$ images, $s(x_i) \leq 1$, $q(x_i) = 1/N$, 
and $\sum_{i=1}^Ng(x_i)=0$, then the variance of $g(x)$ by sampling $M$ images out of 
$N$ images is,
\begin{equation}
\begin{split}
\mathrm{Var}_{q(x)}\bigg[\frac{g(x)p(x)}{q(x)}\bigg] = &  \sum_{i=1}^N\bigg[\frac{g(x_i)p(x_i)}{s(x_i)/N}\bigg]^2\frac{s(x_i)}{N} \\
& - \bigg(\sum_{i=1}^N\bigg[\frac{g(x_i)p(x_i)}{s(x_i)/N}\bigg]\frac{s(x_i)}{N}\bigg)^2 \\
= & \sum_{i=1}^N\bigg[\frac{g(x_i)/N}{s(x_i)/N}\bigg]^2\frac{s(x_i)}{N} \\
& - \bigg(\sum_{i=1}^Ng(x_i)/N\bigg)^2 \\
= & \sum_{i=1}^N\frac{g^2(x_i)}{s(x_i)*N}
\end{split}\label{var2}
\end{equation}
The efficiency is defined as the ratio between ~\ref{var1} and ~\ref{var2}, which is,
\begin{equation}
\begin{split}
R = \frac{\sum_{i=1}^Ng^2(x_i)}{\sum_{i=1}^Ng^2(x_i)/s(x_i)}
\end{split}
\end{equation}
which is the same as the efficiency (relative variance) defined in the main paper. Obviously,
since $s(x_i) \leq 1$, this ratio will always be no larger than 1. If we sample all the data, which
means $s(x_i) = 1$, then we can obtain a sampling efficiency of 1. To simplify the calculation of $R$,
We can further express $\sum_{i=1}^Ng^2(x_i)/s(x_i)$ as,
\begin{equation}
\begin{split}
\sum_{i=1}^N\frac{g^2(x_i)}{s(x_i)} & = \sum_{j=1}^k\frac{\sum_{i=1}^N|g(x_i)|}{M|g(x_j)|}|g(x_j)|^2 + \sum_{j=k+1}^N|g(x_j)|^2 \\
& = \frac{\sum_{i=1}^N|g(x_i)|}{M}(\sum_{j=1}^k|g(x_j)|) + \sum_{j=k+1}^N|g(x_j)|^2 \\
\end{split}
\end{equation}
where $s(x_1)$, $s(x_2)$, $\cdots$, $s(x_k)$ are smaller than 1 and $s(x_{k+1})$, $\cdots$,
$s(x_N)$ are equal to 1.

\end{appendices}
\end{document}